\documentclass[conference]{IEEEtran}

\usepackage{cite}

\usepackage[acronym]{glossaries} 
\newacronym{ai}{AI}{artificial intelligence}
\newacronym{api}{API}{application programing interface}
\newacronym{euxfel}{EuXFEL}{European XFEL}
\newacronym{ide}{IDE}{integrated development environment}
\newacronym{llm}{LLM}{large language model}
\newacronym{mwu}{MWU}{Mann-Whitney U}
\newacronym{ueq}{UEQ}{User Experience Questionnaire}
\newacronym{vscode}{VS Code}{Visual Studio Code}

\usepackage{csquotes}

\usepackage{pbalance}

\usepackage{listings}

\lstdefinestyle{listingstyle}{
    basicstyle=\ttfamily\footnotesize,        
    breaklines=true,       
    frame=lines,
    numbers=none
}
\usepackage{xcolor}
\definecolor{light-gray}{gray}{0.95}

\usepackage{makecell}

\usepackage{multirow}

\usepackage{subcaption}

\usepackage{mdframed}
\usepackage[breakable, many]{tcolorbox} 
\newtcolorbox{answerbox}{
    breakable,
    boxrule = 0pt, 
    leftrule = 2pt,
}

\usepackage{url}

\usepackage{graphicx}

\usepackage{orcidlink}

\hypersetup{
colorlinks=true,
linkcolor=black,
citecolor=black,     
urlcolor=black
}

\begin{document}

\title{Can Developers Prompt? A Controlled Experiment for Code Documentation Generation}

\author{
\IEEEauthorblockN{
Hans-Alexander Kruse \orcidlink{0009-0006-5568-7841},
Tim Puhlfürß \orcidlink{0000-0001-8421-8071},
Walid Maalej \orcidlink{0000-0002-6899-4393}
}
\IEEEauthorblockA{
\textit{Universität Hamburg}\\  
Hamburg, Germany\\
hans-alexander.kruse@studium.uni-hamburg.de,
tim.puhlfuerss@uni-hamburg.de,
walid.maalej@uni-hamburg.de}
}

%\author{
%\IEEEauthorblockN{Anonymized authors}
%\IEEEauthorblockA{\textit{Anonymized organizations} \\  
%Anonymized locations \\
%Anonymized contact information}
%}

\maketitle

\begin{abstract}
% Context
\Glspl{llm} bear great potential for automating tedious development tasks such as creating and maintaining code documentation.
% Objective
However, it is unclear to what extent developers can effectively prompt \glspl{llm} to create concise and useful documentation.
% Methodology
We report on a controlled experiment with 20 professionals and 30 computer science students tasked with code documentation generation for two Python functions.
The experimental group freely entered ad-hoc prompts in a \textit{ChatGPT}-like extension of Visual Studio Code, while the control group executed a predefined few-shot prompt.
% Results
Our results reveal that professionals and students were unaware of or unable to apply prompt engineering techniques.
Especially students perceived the documentation produced from ad-hoc prompts as significantly less readable, less concise, and less helpful than documentation from prepared prompts.
Some professionals produced higher quality documentation by just including the keyword \textit{Docstring} in their ad-hoc prompts.
While students desired more support in formulating prompts, professionals appreciated the flexibility of ad-hoc prompting.
Participants in both groups rarely assessed the output as perfect.
Instead, they understood the tools as support to iteratively refine the documentation.
% Conclusion
Further research is needed to understand which prompting skills and preferences developers have and which support they need for certain tasks.
\end{abstract}


\maketitle

% Main text.
% Advantage of storing the main text in a separate file: You can easily transfer it to another Latex project with a different template.
%---------------------
% INTRO
%---------------------
\section{Introduction}
\label{sec:intro}

% Code documentation as a demanding task
Developers often overlook or ignore software documentation and generally assign it a low priority \cite{aghajani-icse-2019}. 
Yet, carefully documenting code is an essential task in software engineering. 
Up-to-date and high-quality documentation facilitates program comprehension \cite{roehm-icse-2012}, accelerates developer onboarding \cite{steinmacher-software-2019, stanik-icsme-2018}, and mitigates technical debt \cite{zampetti-emse-2021}. 
Software documentation is also central to software maintenance, as documentation often requires updates and should evolve with software  \cite{aghajani-icse-2020}.

% Documentation generation via LLMs
Researchers and tool vendors have thus investigated different ways to automate documentation \cite{rai-tist-2022}. 
In particular, \glspl{llm} designed to model and generate human language \cite{bubeck-arxiv-2023} hold great potential in automating documentation tasks \cite{ebert-software-2023}.
Among popular \gls{llm} use cases, code summarization has recently gained much attention \cite{tian-arxiv-2023}.
However, the output of \glspl{llm} usually depends on the user input, called \textit{prompt}. 
Slightly changing the prompt can lead to different results regarding conciseness, language style, and content of generated text \cite{wermelinger-sigcse-2023}. 

% Prompt engineering
Several prompt engineering techniques have emerged to optimize interactions with \glspl{llm} \cite{liu-compsurv-2023, schulhoff-arxiv-2024, white-arxiv-2023, openai-online-2024}.
Few-shot prompting is one of the most prominent techniques and comprises adding example outputs to a predefined prompt to specify output requirements \cite{brown-neurips-2020, logan-acl-2022}.
This technique allows users to optimize \gls{llm} responses while minimizing the number of input messages sent to the model \cite{white-arxiv-2023, tian-arxiv-2023}.

% Need for a user study + RQs
Previous research has primarily focused on benchmarking the performance of various models and prompt engineering techniques for generating source code documentation \cite{rai-tist-2022}.
Recently, Ahmed and Devanbu concluded that common prompt engineering techniques perform better than ad-hoc prompting according to popular benchmark metrics \cite{ahmed-ase-2022}.
However, studies focusing on the perspectives of developers when using \glspl{llm}, e.g., to generate documentation, are still sparse.
In fact, a recent study showed that common metrics to evaluate \glspl{llm} do not align with human evaluation to assess the quality of generated documentation \cite{hu-tosem-2022}.
Hence, it remains uncertain whether prompt engineering techniques meet developers' requirements for generating and using code documentation.
Moreover, it is unclear whether developers prefer a flexible, iterative interaction using chatbot-like ad-hoc prompting or a rather transaction-like execution of predefined prompts.

This study takes a first step towards filling this gap, focusing on two research questions:

\begin{itemize}[\IEEEsetlabelwidth{RQ}]
\item[\textbf{RQ1:}] How well can developers prompt \glspl{llm} to generate code documentation compared to a predefined prompt?
\item[\textbf{RQ2:}] How is the developer experience for ad-hoc prompting compared to executing predefined prompts?
\end{itemize}

% Methodology + Results + Contribution
To answer these questions, we conducted a randomized controlled experiment with 20 professional developers and 30 computer science students.
The experiment involved generating code documentation using an \gls{llm}-powered \gls{ide}.
The first group utilized a \textit{\gls{vscode}} extension based on GPT-4, enabling participants to enter ad-hoc prompts for selected code.
The second group used a similar \gls{vscode} extension that executed a predefined few-shot prompt to generate code documentation via GPT-4 with a single click.

\begin{figure*}
\centering
\includegraphics[width=\textwidth]{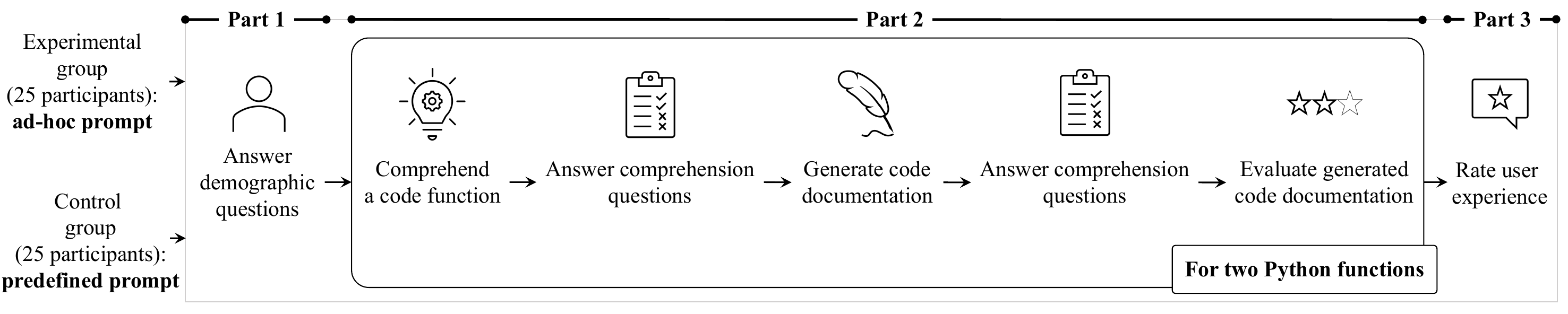}
\caption{Tasks of the between-subject experiment. Participants conducted the second part twice, with different code functions.}
\label{fig:methdology}
\end{figure*}

Overall, we observed that developers with less experience require more guidance in phrasing an ad-hoc prompt, whereas more experienced developers can generate fairly good documentation by including specific keywords like \textquote{Docstring} in the prompts.
Overall, participants preferred transactional or guided \gls{llm} interactions to create code documentation while enjoying the flexibility of ad-hoc and iterative prompting.
Our results show that predefined prompts deliver code documentation of significantly higher quality than ad-hoc prompts, especially due to a consistent documentation format.

We present the design of our study in Section \ref{sec:method} and report on the results in Section \ref{sec:re_compare} (RQ1) and Section \ref{sec:re_ue} (RQ2). 
We then discuss the implication of our findings in Section \ref{sec:disc} and potential threats to validity in Section \ref{sec:threats}. 
Finally, Section \ref{sec:rw} summarizes related work, and Section \ref{sec:conc} concludes the paper.
We share our experiment results, including the \gls{ide} extension, in our replication package \cite{kruse-zenodo-2024}.

%---------------------
% METHODOLOGY
%---------------------
\section{Methodology}
\label{sec:method}

\subsection{Experimental Design}
\label{sec:me_te_ed}

% Overview + controlled variables
We designed a controlled experiment in a laboratory setting to compare developer usage of and experience with two \gls{vscode} extensions employing different prompts for code documentation generation.
We followed a between-subject design, with one participant testing only one of both tools to avoid learning effects \cite{wohlin-springer-2012}.
The ad-hoc prompt group created prompts in an already available GPT-4-powered \gls{ide} extension, while the predefined prompt group interacted with our extension.
We randomly assigned participants to the groups, with adjustments made to balance the occupations. 
Participants were unaware of their group.

% Experiment setup
We offered offline and online versions of the experiment.
In the offline version, participants used our computer with \gls{vscode} and the digital questionnaire displayed on two screens.
In the online version, we utilized \textit{Zoom} for video conferencing, sharing a questionnaire hyperlink, and screen-sharing our \gls{vscode} window with external input control.
Participants employed \textit{Thinking Aloud} to express thoughts during the task \cite{alhadreti-chi-2019}.
To ensure a calm environment, we put a participant in a room empty of people.
The questionnaire included questions about the code and the generated documentation.
Additionally, we collected the ad-hoc prompts for a later comparison with the predefined prompt.

% Experiment tasks (see Figure 1)
The experiment comprised three parts, as shown in Figure \ref{fig:methdology}.
In the \textbf{first part}, participants answered questions about occupation, as well as Python and \gls{vscode} experience to assess their ability to understand complex functions and use the \gls{ide}.
Furthermore, we could check whether these variables have any effect on the observed behavior.

The \textbf{second part} focused on code comprehension and documentation generation.
It consisted of two rounds, each involving a pre-selected Python function (Listing \ref{lst:exp_fct}).
Per round, participants (1) first attempted to comprehend the code without any documentation.
This enabled us to analyze the effect of the \gls{llm}-generated documentation on understanding the code functions.
Subsequently, (2) participants rated their comprehension on a scale from 1 (very low) to 5 (very high), and answered True/False questions about the code to further check their understanding (Table \ref{tab:compr_questions}).
Participants could also state that they had insufficient information to answer a question.
Furthermore, (3) participants manually created a function comment to externalize their understanding and better assess the quality of the comment subsequently generated by the tool.
To guide the participants and shorten the completion time, we provided the participants with a comment template adhering to guidelines for clean Python code \cite{kapil-apress-2019} and instructed them to add only the function description without the explanations of parameters.
Participants then used the respective \gls{llm}-powered \gls{ide} extension to generate code documentation.
Afterward, (4) as documentation should serve to facilitate code comprehension, they studied the generated comment and revisited the True/False questions.
This step provided insights if the documentation supported or even hindered the comprehension.
Finally, (5) they rated the generated documentation on six quality dimensions based on the human evaluation metrics by Hu et al.~\cite{hu-tosem-2022}.

The documentation quality dimensions each provided a question and a five-point answer scale, addressing grammatical correctness, readability, missing relevant information, unnecessary information presence, usefulness for developers, and helpfulness for code comprehension. 
We made minor adjustments to the original answers of two dimensions as we found them overloaded for our goal.
In particular, the answers for grammatical correctness originally also included a rating of fluency, and the original readability answers also concerned the comment's understandability.
The lowest point of each scale represents the negative pole, like \textquote{very low readability}, while the highest point represents the positive pole, like \textquote{very low amount of unnecessary information}.

In the \textbf{third part} of the experiment, participants assessed the usability of the respective \gls{ide} extension.
They completed the standardized \gls{ueq} with 26 items, each associated with a seven-point answer scale and one of six usability categories \cite{laugwitz-hci-2008}.
Finally, participants provided comments on tool strengths and areas for improvement in two free-text fields.

\subsection{Ethical Considerations}
\label{sec:ethics}

We followed the standard procedure of the ethics committee of our department for assessing and mitigating risks regarding ethics and data processing.
Upon welcoming participants, we explained the privacy policy, study purpose, upcoming tasks, and data processing procedures.
We designed the study to last for approx.~30 minutes per participant to prevent exhaustion and instructed participants not to seek perfect solutions to reduce stress. 
We offered clarifications on tasks and questionnaire content without providing solutions.
We minimized observational bias by facing away while participants answered the questionnaire and also encouraged negative feedback \cite{macefield-jus-2007}.
To express our gratitude for the participation, we raffled vouchers among participants.
We pseudonymized all published data as far as possible while maintaining our study objective.

\subsection{Technical Setup}
\label{sec:me_tc}

\begin{lstlisting}[float, caption={Python functions that participants had to comprehend and for which they generated documentation.}, label={lst:exp_fct}]{list=no}
Function 1:
def string_from_vector_bool(data):
    return ",".join(str(int(i)) for i in data)
----------------------------------------------------
Function 2:
def _parse_date(date):
    if date is None:
        date = Timestamp()
    if isinstance(date, Timestamp):
        date = date.toLocal()
    d = dateutil.parser.parse(date)
    if d.tzinfo is None:
        d = d.replace(tzinfo=dateutil.tz.tzlocal())
    return d.astimezone(dateutil.tz.tzutc())
    .replace(tzinfo=None).isoformat()
\end{lstlisting}

\begin{lstlisting}[float, caption={Predefined few-shot prompt. Few-shots 2 and 3 are comparable to 1 and included in the replication package \cite{kruse-zenodo-2024}.}, label={lst:const_prompt}, xleftmargin=0pt]{list=no}
For the following prompt, take into account these 
three input/output pairs of functions and 
corresponding appropriate comments:

Function 1:
def elapsed_tid(cls, reference, new):
    time_difference = new.toTimestamp() - reference.toTimestamp()
    return np.int64(time_difference * 1.0e6 // cls._period)
Comment 1:
"""
    Calculate the elapsed trainId between reference and newest timestamp

    :param reference: the reference timestamp
    :param new: the new timestamp

    :type reference: Timestamp
    :type new: Timestamp

    :returns: elapsed trainId's between reference and new timestamp
"""
Function 2:
...

Generate a comment for the following function: 
{FUNCTION_CONTENT}. Fill in this template: 
{TEMPLATE}. Adhere to the appropriate comment 
syntax for multi-line comments.
\end{lstlisting}

\subsubsection{Selection of Experiment Tasks}
\label{sec:me_te_css}

We selected the Python source code for the experiment tasks from the open-source control system \textit{Karabo} \cite{euxfel-online-2024}.
Karabo is maintained by the scientific facility \textit{\gls{euxfel}}, whose developers partially participated in the experiment.
By choosing this project and involving project insiders, we could assess how their programming and domain experience supports their code comprehension and documentation rating, especially in comparison to outsiders.

Two authors conducted a thorough screening of available Karabo functions and identified six candidates.
All six were utility functions, which we deemed comprehensible for experiment participants not associated with \gls{euxfel}.
Our selection criteria also included code documentation quality aspects such as conciseness, completeness, and usefulness especially relevant for the few-shot examples \cite{rani-jss-2023}.
We independently analyzed the candidates and categorized them into the complexity levels \textit{easy}, \textit{medium}, and \textit{hard}.
Afterward, we resolved categorization conflicts by discussion.

We selected one easy-, and one medium-rated function for the tasks of the experiment (Listing \ref{lst:exp_fct}).
The first function converts a boolean vector into a comma-separated string.
We considered the code easy to understand as only standard Python operations were used within this short function.
The second function converts a given date to the \textit{ISO} format.
The complexity of this function lies in the usage of multiple classes and functions with abbreviated identifiers, and the lack of inline comments.
Hence, participants must make assumptions based on identifier names and usage.

Initially, we also selected a third function rated hard but removed it after a pilot study with three software engineering researchers, who reported difficulties with this function and far exceeded the time limit.
While we acknowledge that \glspl{llm} might require more sophisticated prompts and context information to generate appropriate documentation for more complex code, our user study focused on how differently skilled developers can and prefer to interact with an \gls{llm} for the creative process of documentation generation.
Therefore, within the experiment's time limit, all participants should be able to phrase ad-hoc prompts that generate appropriate comments for these functions.

\subsubsection{LLM Selection and Prompt Construction}
\label{sec:me_tc_ms}

Research and industry have introduced multiple \glspl{llm} with individual strengths and limitations. 
For our study, we focused solely on the GPT-4 model by OpenAI (version \textit{gpt-4-0613}). 
We did not aim to benchmark different models but rather to study developers prompting skills particularly for code documentation generation.
OpenAI's models have been prominent in recent research on documentation generation \cite{hou-arxiv-2023}. 
Furthermore, during our study period, GPT-4 demonstrated general superiority over other models \cite{bubeck-arxiv-2023, hu-tosem-2022}.

We defined the predefined few-shot prompt (Listing \ref{lst:const_prompt}) based on prompt engineering guidelines \cite{white-arxiv-2023, sun-arxiv-2023, openai-online-2024} and tested it in a pilot study.
Given GPT-4's limited context window at the time of the study, we applied a three-shot approach to specify the required output format.
We chose three distinct function-comment pairs from the remaining previously selected Karabo code candidates.
Hence, these functions were from the same domain and of similar complexity as the two used in the experiment.
They adhered to a structured documentation format with precise and concise content.
We instructed GPT-4 to generate a comment for the selected source code that we passed as input (\textit{FUNCTION\_CONTENT}).
To further enforce the clean code format \cite{kapil-apress-2019}, we explicitly defined an output template within the prompt (\textit{TEMPLATE}) and requested adherence to Python's syntax for multi-line comments (\textit{Docstrings}).
Listing \ref{lst:const_prompt} only contains example 1 of the few-shot prompt. 
Examples 2 and 3 are available in our replication package \cite{kruse-zenodo-2024}.
We acknowledge that the prompt can be further optimized and generalized to multiple programming languages and application domains.
We considered its performance and complexity sufficient for our research goal.

\subsubsection{Tool of the Ad-hoc Prompt Group}
\label{sec:me_te_bts}

\begin{figure}
\centering
\begin{subfigure}{\columnwidth}
\centering
\includegraphics[width=\columnwidth]{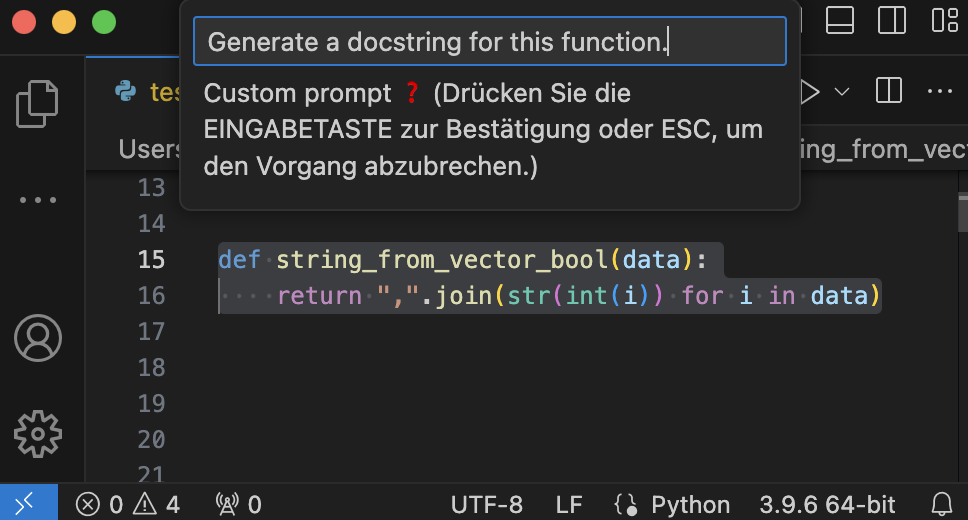}
\caption{Input field of the ad-hoc prompt tool.}
\label{fig:adhoc_tool}
\end{subfigure}
\begin{subfigure}{\columnwidth}
\includegraphics[width=\columnwidth]{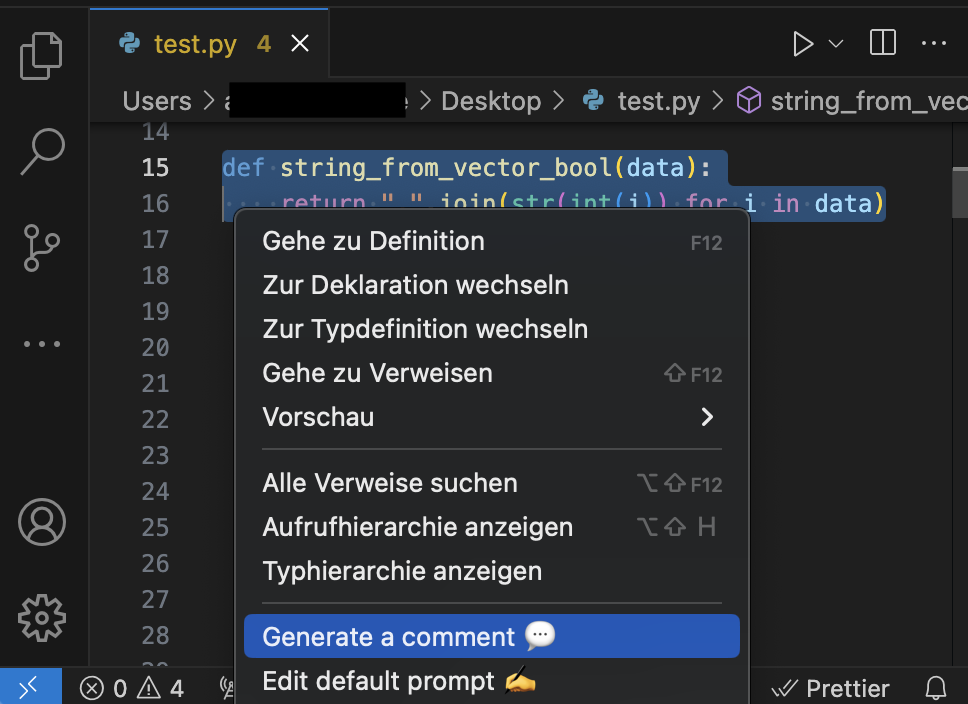}
\caption{\textquote{Generate a comment} action of the predefined prompt tool.}
\label{fig:predefined_tool}
\end{subfigure}
\caption{\gls{vscode} extensions applied in the experiment.}
\end{figure}

We focused on \gls{ide} extensions as developers usually use \glspl{ide} for generating code documentation.
We chose \gls{vscode} for its extensive customization capabilities through various extensions maintained by an open-source community. 
During our study, the extension marketplace already offered multiple ChatGPT-like tools that we examined.
Additionally, \gls{vscode} supports multiple programming languages, including Python.

We selected the ad-hoc prompt tool based on two criteria.
First, it needed to provide a user interface similar to the ChatGPT interface, allowing users to engage in a conversation with the software. 
This approach involves at least one self-written message to the model and copying and pasting the generated comment from the chat interface to the code.
This criterion was crucial as we aimed to compare the user experience for this classic Q\&A-based interaction with the one for the predefined prompt tool, which generates and inserts code documentation with a button click.
Second, the tool had to support and behave like the original GPT-4 \gls{api} to handle any confounding factors related to the use of different \gls{llm} models.

After a thorough testing of available extensions, we opted for \textit{ChatGPT} by \textit{zhang-renyang} \cite{renyang-online-2024} in version \textit{1.6.62}. 
This extension met all specified requirements.
It offers a small prompt field (Figure \ref{fig:adhoc_tool}) and a larger ChatGPT-like chat window within the \gls{ide}.
Our test showed that it mimics GPT-4.
It was one of the most popular GPT extensions in the marketplace during our study period.

\subsubsection{Tool of the Predefined Prompt Group}
\label{sec:me_tc_ti}

% Tool design goals
Our design objective for the predefined prompt group was to minimize the steps required for developers to generate a high-quality Docstring comment for a specific code section. 
We determined that marking the code within the \gls{ide}'s code editor and selecting the \textquote{Generate a comment} action from the context menu should suffice, requiring three clicks to generate documentation (Figure \ref{fig:predefined_tool}).
Alternatively, the developer can choose this action through \gls{vscode}'s command field. 
Hence, inexperienced and experienced \gls{vscode} users can use the primary feature of the tool with their preferred workflow.
A secondary feature allows developers to customize the predefined prompt via the \textquote{Edit default prompt} action, enabling them to tailor the tool output to their preferred format.

% Implementation details
We followed the \gls{vscode} guidelines to create the \textit{TypeScript}-based extension \cite{microsoft-online-2024} and utilized the \textit{axios} package to implement calls to the OpenAI \gls{api}.
Based on the input prompt, the \gls{api} returns a code comment generated by GPT-4.
The extension automatically inserts this documentation above the previously selected code.
To set up the extension, developers must enter their OpenAI \gls{api} token into a configuration field.
This token is stored locally on the developer's machine.
For the experiment, we pre-configured the tool with our token.

% GPT and VS Code versions
We used GPT-4 version \textit{gpt-4-0613} for both tools. 
Generating code documentation with this model took, on average, four seconds.
The \gls{vscode} version was \textit{1.85}.
The predefined prompt tool is included in our replication package \cite{kruse-zenodo-2024} and also available in the \gls{vscode} marketplace \cite{redevtools-online-2024}.

\subsection{Experiment Execution and Participants}
\label{sec:me_te_eee}

% Software versions, Demographics of the participants, study duration, mean length of the experiments
The experiment took place in October and November 2023, involving 50 participants: 25 per ad-hoc and predefined prompt group. 
20 participants were professionals, including 12 software engineers, four data scientists, and four with other software development roles. 17 worked in the domains of scientific computing and three in finance.
These participants took the \textbf{\textit{expert}} perspective due to their professional experience with Python programming and, eventually, the application domain.
The remaining 30 participants were students in computer science or related subjects, taking the \textbf{\textit{newcomer}} perspective.
We recruited participants through personal contacts in academia and industry to ensure that the participants had at least some experience with Python and \gls{vscode}. 
We conducted 33 experiments online and 17 offline at the \gls{euxfel} campus in Schenefeld, Germany, with the first author leading the experiment and the second author assisting.

On a scale from 1 (very low) to 5 (very high), the mean value and standard deviation of the Python experience among professionals were 3.7 (1.06) in the ad-hoc and 4.5 (0.71) in the predefined prompt group.
The values among students were notably lower, with 2.53 (0.83) in the ad-hoc and 2.47 (1.19) in the predefined prompt group.
This difference confirms the expert/newcomer setting.

All participants were familiar with \gls{vscode}, which was important to mitigate biases due to the tool setup.
Mean values and standard deviations in the ad-hoc/predefined prompt group among professionals were 2.9 (0.99) and 3.5 (1.18); among students they were 3.07 (1.03) and 3.07 (1.28).

At the beginning of each session, we trained each participant in using the respective tool to mitigate learning bias.
We introduced them to the tool features, generated documentation for an example function, and answered tool-related questions.
Afterward, we started the experiment.
Due to manually writing prompts, ad-hoc prompt group participants completed the experiment after 23:53 minutes on average, while predefined prompt group participants required only 19:58 minutes.

\subsection{Data Analysis}
\label{sec:me_te_ex}

% Descriptive and inferential statistics
Our experiment comprised quantitative and qualitative data analyses.
We report totals, mean values, and standard deviations where applicable.
Furthermore, we employed inferential statistics to test for significant differences between the groups' ratings of the six documentation quality dimensions. 
Depending on the statistical assumptions fulfilled by the respective variables, we utilized the parametric \textit{Welch's t test} or the non-parametric \textit{\gls{mwu} test} \cite{derrick-tqmp-2016, nachar-tqmp-2008}.

The assumptions for both tests were that the collected data points were quantitative and independent.
The first assumption was fulfilled as we provided a five-point answer scale for each question related to a quality dimension.
The second assumption was fulfilled as we conducted the experiment with each participant individually, and we asked participants to not share information with others to avoid external influences.
The third assumption of the Welch's t test is a normal distribution of data points, which we tested with the \textit{Shapiro-Wilk test} \cite{mohd-josmaa-2011}.
If a normal distribution was not met for a specific quality dimension, we applied the \gls{mwu} test instead.
Besides reporting the p-values for statistical significance, we also report the effect sizes to indicate practical significance.
We particularly applied \textit{Cliff's $\delta$} \cite{cohen-book-1988}, which describes the overlap of two groups of ordinal numbers.
Values between 0.0 and 0.8 indicate a low to medium effect size, whereas values from 0.8 represent a large effect size, meaning a large difference between both groups.

% UEQ
To quantify the experience of developers using the experiment tools, we utilized the 26 seven-point scales of the \gls{ueq} \cite{laugwitz-hci-2008} and its official data analysis tool \cite{ueq-online-2024}.
This tool provides descriptive (mean values and standard deviations) and inferential statistics (t test) to compare user experience ratings for the two studied \gls{ide} extensions.

Each of the 26 scales is related to one of six quality categories.
The first category, \textit{Attractiveness}, represents the pure valence of and overall satisfaction with a tool.
\textit{Efficiency}, \textit{Perspicuity}, and \textit{Dependability} are the pragmatic, goal-directed categories, and \textit{Stimulation} and \textit{Novelty} are hedonistic, not goal-directed categories.
We chose Efficiency, Perspicuity, Dependability, and Novelty to measure tool-related experience, as these categories focus on tool interactions.

To evaluate participants' assessment of the quality of generated documentation, we used the \textit{Stimulation} category as it focuses on the output of a tool.
We enhanced this analysis with qualitative insights made by manually analyzing participants' free-text answers to the questions \textquote{What did you like most about the tool?} and  \textquote{What would you change about the tool?}, placed at the end of the questionnaire. 
Two authors independently conducted \textit{Open Coding} for all answers and discussed their codes to achieve consensus \cite{cascio-fm-2019}.
Furthermore, we report relevant remarks expressed by participants during the experiment.

\begin{table}
\centering
\caption{Frequency of successful ad-hoc prompts for both tasks, subdivided by occupations.}
\begin{tabular}{lccc}
\hline
\textbf{Prompt pattern} & \textbf{All} & \textbf{Students} & \textbf{Professionals} \\ 
\hline
Write a comment & 34 & 24 & 10 \\
Write a Docstring & 12 & 3 & 9 \\
Explain the function in a comment & 2 & 2 & 0 \\
Write a Python-conform description & 1 & 1 & 0 \\
Summarize the function in a comment & 1 & 0 & 1 \\
\hline
\end{tabular}
\label{tab:prompt_pat}
\end{table}

\begin{table*}
\centering
\caption{Ratings of the generated documentation along six quality dimensions (mean values and standard deviations of 5-point-scales) in ad-hoc and predefined prompt groups; for all participants, students, and professionals; for code functions 1 and 2. 
Higher ratings always imply better quality. 
Every third line presents the p-value of the \gls{mwu} test and Cohen's Delta.}
\label{tab:ratings_groups}
\begin{tabular}{ccccccccc}
% All
\hline
\textbf{Group} & \textbf{Subgroup} & \textbf{\makecell{Code\\Function}} & \textbf{\makecell{Grammatical\\Correctness}} & \textbf{Readability} & \textbf{\makecell{Missing\\Information}} & \textbf{\makecell{Unnecessary\\Information}} & \textbf{Usefulness} & \textbf{Helpfulness} \\
\hline
\textbf{Ad-hoc} & \textbf{All} & \textbf{1} & 4.88 (0.33) & 4.12 (0.88) & 4.64 (0.76) & 3.28 (1.21) & 4.12 (0.83) & 4.52 (0.59) \\
\textbf{Predefined} & \textbf{All} & \textbf{1} & 4.88 (0.33) & 4.72 (0.54) & 4.36 (1.11) & 4.32 (1.25) & 4.44 (1.04) & 4.32 (1.07) \\
\textbf{p-val. \(|\) $\delta$} & \textbf{All} & \textbf{1} & 1.0 \(|\) 0.0 & \textbf{0.005 \(|\) 0.82} & 0.35 \(|\) 0.29 & \textbf{0.0009 \(|\) 0.85} & 0.08 \(|\) 0.3 & 0.91 \(|\) 0.23 \\
\hline
\textbf{Ad-hoc} & \textbf{All} & \textbf{2} & 4.48 (0.92) & 3.52 (1.39) & 3.92 (1.29) & 2.84 (1.60) & 3.52 (1.39) & 4.0 (1.08) \\
\textbf{Predefined} & \textbf{All} & \textbf{2} & 4.76 (0.66) & 4.68 (0.48) & 4.16 (0.85) & 4.68 (0.85) & 4.6 (0.76) & 4.6 (0.82) \\
\textbf{p-val. \(|\) $\delta$} & \textbf{All} & \textbf{2} & 0.12 \(|\) 0.35 & \textbf{0.0004 \(|\) 1.12} & 0.78 \(|\) 0.22 & \textbf{0.00002 \(|\) 1.44} & \textbf{0.002 \(|\) 0.96} & \textbf{0.02 \(|\) 0.63} \\ \hline
% Students
\hline
\textbf{Ad-hoc} & \textbf{Students} & \textbf{1} & 4.86 (0.35) & 4.0 (1.0) & 4.6 (0.63) & 3.33 (1.35) & 3.93 (0.88) & 4.46 (0.64) \\
\textbf{Predefined} & \textbf{Students} & \textbf{1} & 4.93 (0.26) & 4.66 (0.62) & 4.6 (0.83) & 4.86 (0.35) & 4.6 (0.63) & 4.6 (0.63) \\
\textbf{p-val. \(|\) $\delta$} & \textbf{Students} & \textbf{1} & 0.58 \(|\) 0.22 & \textbf{0.03 \(|\) 0.8} & 0.76 \(|\) 0.0 & \textbf{0.0002 \(|\) 1.56} & \textbf{0.03 \(|\) 0.87} & 0.52 \(|\) 0.21 \\
\hline 
\textbf{Ad-hoc} & \textbf{Students} & \textbf{2} & 4.4 (1.06) & 3.2 (1.47) & 3.86 (1.30) & 2.66 (1.68) & 3.13 (1.41) & 3.66 (1.23) \\
\textbf{Predefined} & \textbf{Students} & \textbf{2} & 4.73 (0.80) & 4.6 (0.51) & 4.06 (0.96) & 4.86 (0.35) & 4.73 (0.46) & 4.6 (0.74) \\
\textbf{p-val. \(|\) $\delta$} & \textbf{Students} & \textbf{2} & 0.13 \(|\) 0.36 & \textbf{0.002 \(|\) 1.27} & 0.84 \(|\) 0.17 & \textbf{0.0002 \(|\) 1.82} & \textbf{0.0008 \(|\) 1.53} & \textbf{0.02 \(|\) 0.92} \\
% Professionals
\hline \hline
\textbf{Ad-hoc} & \textbf{Professionals} & \textbf{1} & 4.9 (0.32) & 4.3 (0.67) & 4.7 (0.95) & 3.2 (1.03) & 4.4 (0.70) & 4.6 (0.52) \\
\textbf{Predefined} & \textbf{Professionals} & \textbf{1} & 4.8 (0.42) & 4.8 (0.42) & 4.0 (1.41) & 3.5 (1.65) & 4.1 (1.45) & 3.9 (1.45) \\
\textbf{p-val. \(|\) $\delta$} & \textbf{Professionals} & \textbf{1} & 0.58 \(|\) 0.27 & 0.07 \(|\) \textbf{0.89} & 0.08 \(|\) 0.58 & 0.51 \(|\) 0.22 & 0.97 \(|\) 0.26 & 0.38 \(|\) 0.64 \\
\hline
\textbf{Ad-hoc} & \textbf{Professionals} & \textbf{2} & 4.6 (0.70) & 4.0 (1.15) & 4.0 (1.33) & 3.1 (1.52) & 4.1 (1.20) & 4.5 (0.53) \\
\textbf{Predefined} & \textbf{Professionals} & \textbf{2} & 4.8 (0.42) & 4.8 (0.42) & 4.3 (0.67) & 4.4 (1.26) & 4.4 (1.07) & 4.6 (0.97) \\
\textbf{p-val. \(|\) $\delta$} & \textbf{Professionals} & \textbf{2} & 0.58 \(|\) 0.35 & 0.06 \(|\) \textbf{0.92} & 0.9 \(|\) 0.28 & 0.07 \(|\) \textbf{0.93} & 0.47 \(|\) 0.26 & 0.28 \(|\) 0.13 \\
\hline
\end{tabular}
\end{table*}

%---------------------
% RESULTS - RQ1
%---------------------
\section{Prompting for Code Documentation (RQ1)}
\label{sec:re_compare}

\subsection{Patterns in Ad-hoc Prompts}
\label{sec:re_prompt}

% Failed attempts
The prompts entered by the ad-hoc prompt group varied in detail and keywords. 
About half of the participants had to re-enter a different prompt after their first attempt to generate a code comment.
Among the failed attempts were prompts such as \textquote{Explain the function} and \textquote{Describe what this function does}, which summarized the code in longer prose text.

% Patterns of successful prompts
We categorized the 50 prompts that led to a successful generation of code comments for both code functions into five prompt patterns, listed in Table \ref{tab:prompt_pat}.
These patterns commonly included documentation-specific keywords, i.e., \textit{comment} (34x), \textit{Docstring} (12), and \textit{Python-conform description} (1).
Also, slightly longer prompts that included the comprehension-focused terms \textit{explain} (2) and \textit{summarize} (1) led to appropriate function comments.
Professionals used the Docstring term more often (9x) than students (3x), which is in line with the higher Python experience of most participating professionals.

% Prompt lengths and content
Successful prompts were generally short, with a mean length and standard deviation of 7.1 (2.09) words among professionals and 6.7 (2.62) words among students.
The shortest prompt was \textquote{Generate docstring}.
None of the participants enhanced their prompts with examples or templates.

\subsection{Perceived Quality of Generated Documentation}
\label{sec:re_co_ov}

% Overview of the process and results
Participants rated the documentation generated by the ad-hoc and predefined prompt tools based on six five-point-scale questions.
All participants were aware of the expected format of a Python Docstring, as they were familiar with Python. 
Moreover, before using the tool, they manually created a function comment based on a Docstring template.
Table \ref{tab:ratings_groups} displays the ratings per group, subgroup (student vs. professional), code function, and quality dimension, along with the p-values \cite{derrick-tqmp-2016, nachar-tqmp-2008} and effect sizes \cite{cohen-book-1988}.

% Choice of statistical test
We performed several trials to determine the appropriate statistical test (see section \ref{sec:me_te_ex} for details). 
We rejected the null hypothesis of the Shapiro-Wilk test for all groups and dimensions, indicating a non-normal data distribution, likely due to small sample sizes. 
We chose the \gls{mwu} test for all comparisons due to its suitability for non-normally distributed data \cite{nachar-tqmp-2008}.
The null hypothesis of this test indicates no statistically significant difference between the mean values of both groups for a specific dimension. 
We rejected this null hypothesis if the p-value was lower than 0.05.
We noted that in all cases when the null hypothesis was rejected, the effect size $\delta$ was large (0.8 or higher), indicating practical significance \cite{cohen-book-1988}.

\begin{table*}
\centering
\caption{Count of correct answers for comprehension questions (Q1, Q2) for code functions F1 and F2 in the ad-hoc and predefined prompt groups: before/after using the tool. 
Colors indicate if correct answers increased after using the tool.}
\begin{tabular}{l l c c c c c c}
\hline
& & \multicolumn{2}{c}{\textbf{All}} & \multicolumn{2}{c}{\textbf{Students}} & \multicolumn{2}{c}{\textbf{Professionals}}\\
\textbf{ID} & \textbf{Question} & \textbf{Ad-hoc} & \textbf{Predef.} & \textbf{Ad-hoc} & \textbf{Predef.} & \textbf{Ad-hoc} & \textbf{Predef.}\\ \hline
\textbf{F1 Q1} & \textbf{Does the function return a list of boolean values as its output?} & 19/\textcolor{teal}{22} & 23/\textcolor{teal}{24} & 11/\textcolor{teal}{14} & 13/\textcolor{teal}{15} & 8/8 & 10/\textcolor{red}{9}\\
\textbf{F1 Q2} & \textbf{Does the function take an iterable as its input?} & 23/\textcolor{red}{21} & 24/\textcolor{red}{20} & 14/\textcolor{red}{13} & 14/\textcolor{red}{12} & 9/\textcolor{red}{8} & 10/\textcolor{red}{8}\\ \hline
\textbf{F2 Q1} & \textbf{Does the function modify the input date's time zone to UTC?} & \textbf{10/\textcolor{teal}{22}} & 14/\textcolor{teal}{18} & \textbf{5/\textcolor{teal}{13}} & 8/\textcolor{teal}{11} & 5/\textcolor{teal}{9} & 6/\textcolor{teal}{7}\\
\textbf{F2 Q2} & \textbf{Does the function parse JSON data into a dictionary structure?} & 15/\textcolor{teal}{21} & 13/\textcolor{teal}{21} & 7/\textcolor{teal}{12} & \textbf{5/\textcolor{teal}{14}} & 8/\textcolor{teal}{9} & 8/\textcolor{red}{7}\\ 
\hline
\end{tabular}
\label{tab:compr_questions}
\end{table*}

% Comparing All Participants
Comparing \textbf{all participants} in the ad-hoc and predefined prompt groups, we observed several statistically significant differences.
For function 1, we found significant differences in the dimensions of Readability (4.12/4.72) and Unnecessary Information (3.28/4.32).
The predefined prompt tool consistently provided concise Docstrings that participants perceived as more readable and contained less non-informative content compared to the output of the ad-hoc prompt tool.
For function 2, significant differences were found in four dimensions, with the predefined prompt tool consistently receiving higher ratings than the ad-hoc prompt tool.
The differences in Readability (3.52/4.68) and Unnecessary Information (2.84/4.68) reinforced the conclusions drawn from function 1.

We attribute the ratings of Usefulness (3.52/4.6) to the often excessive output length of the ad-hoc prompt tool.
Participants noted that longer, prose documentation could impede development workflows, as these comments require more time to read while providing little additional information.
Helpfulness was rated high in both groups (4.0/4.6), with some participants suggesting that the explanations provided by both tools were either too complicated or lacked the required level of detail.
Across both groups, participants rated Grammatical Correctness and Missing Information very high.

% Students vs. Professionals
When we compared documentation ratings between students and professionals, we observed similar statistically significant differences in the ad-hoc and predefined prompt groups for students, while professionals did not exhibit significant differences.
Among \textbf{students}, we observed significant differences for function 1 in the dimensions Readability (4.0/4.66), Unnecessary Information (3.33/4.86), and Usefulness (3.93/4.6).
The different expectations regarding the documentation content contributed to the disparity in Usefulness, with some students preferring the longer summarizations by the ad-hoc prompt tool, while most others expected a structured comment as taught in university.

For function 2, students in the predefined prompt group consistently provided significantly better ratings for Readability (3.2/4.6), Unnecessary Information (2.66/4.86), Usefulness (3.13/4.73), and Helpfulness (3.66/4.6) compared to students in the ad-hoc prompt group.
Unnecessary Information had the highest differences, likely due to the unstructured and lengthy comments generated via the ad-hoc prompts.

Among the \textbf{professionals}, we found no statistically significant differences between the ad-hoc and predefined prompt groups.
However, three dimensions showed large effect sizes ($\geq$ 0.8), indicating a practical significance.
These differences concerned Readability for functions 1 (4.3/4.8) and 2 (4.0/4.8), and Unnecessary Information for function 2 (3.1/4.4).
This indicates that professionals also found the comments of the ad-hoc prompt tool to contain non-informative content, impairing readability.
Conversely, the ad-hoc prompt group provided a more positive rating for the dimension Missing Information (4.7/4.0) for function 1.
Professionals expected more detailed descriptions, which the concise comments by the predefined prompt tool did not provide.
Furthermore, as professionals used the Docstring keyword more often than students in their prompts (Table \ref{tab:prompt_pat}), the ad-hoc prompts in these cases resulted in similar documentation to the predefined prompt tool.

\subsection{Code Comprehension}
\label{sec:re_function}

% Perceived comprehension
We analyzed how the \gls{llm}-generated documentation influenced the comprehension of the code.
Initially, participants rated their perceived comprehension of the respective code function without available documentation, on a scale from 1 (\textquote{not at all}) to 5 (\textquote{very well}).
For function 1, the mean and standard deviation for these ratings were 3.6 (0.96) by the ad-hoc and 3.84 (1.11) by the predefined prompt group, indicating a similar good understanding.
The assessment for function 2 was lower, with 2.4 (0.96) by the ad-hoc and 2.92 (0.81) by the predefined prompt group.
Hence, the predefined prompt group expressed slightly higher confidence in the comprehension than the ad-hoc prompt group, despite similar stated Python experience.
Across both groups and functions, professionals stated a higher comprehension than students, which aligns with their higher Python experience.

% Actual comprehension
We assessed participants' actual comprehension through True/False questions about the code (Table \ref{tab:compr_questions}).
We counted the number of participants who correctly answered these questions before and after generating the documentation. 
Participants could also choose the option \textquote{Not enough information available to answer the question}, which we treated as an incorrect answer.

Before using the tools, most participants in both groups answered the two questions for function 1 correctly (Q1: 19+23; Q2: 23+24), as expected for this short and easy-categorized function.
After using the tool, the correct answers for question 1 increased (Q1: 22+24), especially among students, while the ones for question 2 slightly decreased (Q2: 21+20) for both students and professionals.
Thinking-aloud observations revealed that many participants were confused by the less technical documentation generated by both tools regarding this question, as it did not include the term \textit{iterable}.

The number of correct answers to function 2 before using the tool was lower (Q1: 10+14; Q2: 15+13), reflecting its higher complexity. 
In both groups, correct answers increased after using the tools (Q1: 22+18; Q2: 21+21). 
Students had the highest increases: in the ad-hoc prompt group for Q1 from 5 to 13, and in the predefined prompt group for Q2 from 5 to 14. 
The increase in comprehension among professionals was marginal, except for the correct Q1 answers in the ad-hoc prompt group, which increased from 5 to 9.
This data shows that the availability of documentation helped inexperienced developers to better comprehend the code, while the effect on professionals was mixed.

\begin{answerbox}
\textbf{Answering RQ1}, we conclude that developers were able to generate code documentation using ad-hoc prompts without prompt engineering techniques, often by providing relevant documentation keywords.
However, particularly less experienced students initially did not use such keywords and generated lengthy, less readable, and less useful code explanations instead.

This is reflected in participants rating the quality of generated documentation.
We observed statistically significant differences exclusively among students, who rated the readability, conciseness, usefulness, and helpfulness of comments generated via the predefined few-shot prompt higher than those created via ad-hoc prompts.
Although we observed large effect sizes for some quality dimensions among professionals, this subgroup was generally less critical as they often generated documentation of the required quality and format by using specific keywords.

We also conclude that documentation generated by both tools helped (particularly students) comprehend more complex code.
The generated documentation could also lead to confusion, even for professionals.
\end{answerbox}

\begin{table*}
\centering
\caption{Comparison of developer experience: Ratings of the six UEQ categories (mean values and standard deviations) from the ad-hoc and predefined prompt groups, including the p-values (t test) and Cohen's Delta.}
\begin{tabular}{cccccccc}
\hline
\textbf{Group} & \textbf{Subgroup} & \textbf{Attractiveness} & \textbf{Efficiency} & \textbf{Perspicuity} & \textbf{Dependability} & \textbf{Stimulation} & \textbf{Novelty}\\
\hline
\textbf{Ad-hoc} & \textbf{All} & 1.01 (1.14) & 0.70 (1.19) & 1.25 (0.68) & 1.05 (1.05) & 0.82 (1.13) & 1.10 (1.15)\\ 
\textbf{Predefined} & \textbf{All} & 1.99 (0.63) & 1.98 (0.67) & 1.87 (0.47) & 1.84 (0.79) & 1.62 (0.83) & 1.18 (0.92)\\ 
\textbf{p-val. \(|\) $\delta$} & \textbf{All} & \textbf{0.0006 \(|\) 1.06} & \textbf{0.0000 \(|\) 1.33} & \textbf{0.0005 \(|\) 0.71} & \textbf{0.004 \(|\) 0.85} & \textbf{0.007 \(|\) 0.81} & 0.79 \(|\) 0.08\\
\hline \hline
\textbf{Ad-hoc} & \textbf{Students} & 0.64 (1.14) & 0.42 (1.0) & 1.33 (0.74) & 0.85 (1.15) & 0.6 (1.23) & 0.92 (1.31)\\
\textbf{Predefined} & \textbf{Students} & 1.94 (0.65) & 2.12 (0.68) & 1.98 (0.27) & 1.95 (0.87) & 1.6 (0.88) & 1.18 (0.85)\\  
\textbf{p-val. \(|\) $\delta$} & \textbf{Students} & \textbf{0.0009 \(|\) 1.4} & \textbf{0.0000 \(|\) 1.99} & \textbf{0.005 \(|\) 1.17} & \textbf{0.007 \(|\) 1.08} & \textbf{0.02 \(|\) 0.94} & 0.51 \(|\) 0.24\\
\hline \hline
\textbf{Ad-hoc} & \textbf{Professionals} & 1.55 (0.94) & 1.13 (1.38) & 1.13 (0.58) & 1.35 (0.83) & 1.15 (0.93) & 1.38 (0.84)\\ 
\textbf{Predefined} & \textbf{Professionals} & 2.05 (0.64) & 1.78 (0.64) & 1.7 (0.65) & 1.68 (0.68) & 1.65 (0.78) & 1.18 (1.06)\\ 
\textbf{p-val. \(|\) $\delta$} & \textbf{Professionals} & 0.18 \(|\) 0.62 & 0.2 \(|\) 0.6 & 0.05 \(|\) \textbf{0.93} & 0.35 \(|\) 0.43 & 0.21 \(|\) 0.58 & 0.65 \(|\) 0.21\\
\hline
\end{tabular}
\label{tab:ueq_results}
\end{table*}

%---------------------
% RESULTS - RQ2
%---------------------
\section{Developer Experiences With Prompting (RQ2)}
\label{sec:re_ue}

\subsection{Overall Experience}

% Overview of evaluation results
Participants rated both tools positively in all six categories of the standardized \gls{ueq} \cite{laugwitz-hci-2008}, with values ranging from -3 to +3 (Table \ref{tab:ueq_results}).
However, students found the predefined prompt tool more positive in all user experience categories than the ad-hoc prompt tool, with statistically significant differences and large effect sizes in five categories.
Professionals also rated the predefined prompt tool more positively in five categories, without significant differences but a large effect size in one category.
% Attractiveness
For the pure valence category \textbf{Attractiveness (1.01 ad-hoc / 1.99 predefined prompt group)}, the predefined prompt tool received higher ratings than the ad-hoc prompt tool, with a high difference among students (0.64/1.94). 
This indicates an overall better developer experience with the predefined prompt tool.

\subsection{Tool-Related Experience}

% Efficiency
The most significant difference between the groups was in the category \textbf{Efficiency (0.7/1.98)}, with the predefined prompt tool receiving higher ratings, especially among students (0.42/2.12).
Participants praised the ad-hoc prompt tool for its practicality of not having to open ChatGPT in the browser (two students, one professional), flexibility in selecting relevant code (one professional), and custom prompts (one professional).
They noted issues with copy-pasting the tool output to the code panel (six students, four professionals), multiple tries sometimes required for sufficient results (one student), the need to phrase a prompt (one professional), missing keyboard shortcuts (one professional), and the slow output generation, which is primarily caused by the performance of GPT-4 (one student).
Participants who interacted with the predefined prompt tool also noted the slow response (one student, two professionals), but expressed no further positive or negative comments.

% Perspicuity
The ratings for \textbf{Perspicuity (1.25/1.87)} were higher for the predefined prompt tool, with large effect sizes for students and professionals.
While some participants of the ad-hoc prompt group participants found their tool easy to use (four students, one professional), others reported that they would find it easier to enter the prompt in a separate \gls{ide} panel instead of the pop-up at the top of the screen (one professional) and that the right-click context menu, which listed the tool features besides non-related \gls{ide} features, was obscure to them (one professional).
Multiple predefined prompt group participants also found their tool easy to use (eight students, four professionals), highlighting the ease of generating comments without the need for further prompt input besides the code and a short command (one professional).
Nevertheless, they assessed that the selection of relevant code could be facilitated (one professional).
Overall, this assessment aligns with our observation that participants quickly understood both \gls{ide} extensions but tended to prefer the simplicity of the predefined prompt tool without having to create an effective prompt.

% Dependability
Ratings for \textbf{Dependability (1.05/1.84)} showed strongly diverging views among students (0.85/1.95).
Participants praised the ad-hoc prompt tool for supporting the comment creation workflow (six students, two professionals) and code comprehension (three students, four professionals).
They criticized the lack of prompt templates (three students), that the tool did not always generate code comments (two students, one professional), and the missing opportunity for iterative improvements caused by the limitations of the OpenAI \gls{api} at the time of our study (one student, one professional).
The predefined prompt tool was also noted for facilitating the workflow (three students, one professional) and code comprehension (four students), as well as for its reliability (two professionals).
However, participants suggested adding a progress bar (one professional), previewing generated comments within the function (one professional), enabling the configuration of the required comment content (one student), adding a disclaimer that the output might be incorrect (one student), implementing a linter that alerts when generated comments became outdated (one professional), and using a locally deployed \gls{api} for data security (one professional).
Besides these improvement comments, the predefined prompt group was overall satisfied regarding the Dependability aspects, whereas the ad-hoc prompt group, and especially the students, required more enhanced tool support.

% Novelty 
Both tools achieved similar positive results concerning \textbf{Novelty (1.1/1.18)}, without significant differences.
Interestingly, the ad-hoc prompt tool received a higher rating from professionals (1.38/1.18) for its chat-like interaction within an IDE, which they considered more novel than clicking a button for documentation generation.
Participants expressed no further comments regarding this \gls{ueq} category.

\subsection{Documentation-Related Experience}

% Stimulation
For \textbf{Stimulation (0.82/1.62)}, predefined prompt tool ratings were higher, especially among students (0.6/1.6).
Concerning the output of the ad-hoc prompt tool, participants noted the detailed explanations (two students, two professionals), the mitigation of personal biases (one student), inspirational value (one professional), and overall high quality (three professionals).
Other participants criticized the output for containing unnecessary information (five students, three professionals), lack of detail regarding complex code (three students), and inconsistent format due to using prompts of varying qualities (two students, one professional).
The predefined prompt tool output received positive ratings for its conciseness (four students, one professional), overall high quality (three professionals), and similar structure across all generated comments (one student, one professional).
Improvements were deemed necessary for the parameter descriptions (one student, one professional), line breaks (one professional), consistent punctuation (one student), and lack of detail for complex code (one student, one professional).
These insights show that students were especially unsatisfied with the ad-hoc prompt tool output, while professionals were overall satisfied with both tools.

\begin{answerbox}
\textbf{Answering RQ2}, we conclude that the participants overall preferred a tool that automates code documentation generation with a few clicks while offering options to configure this process.
On average, professionals and students preferred executing predefined prompts over creating ad-hoc prompts.
The simplicity and efficiency of a single button click to receive consistently high-quality documentation were key factors for this assessment.
While the predefined prompt group praised their tool, some missed the flexibility to adjust the documentation depth.
Users of the ad-hoc prompt tool appreciated this flexibility, which often unintentionally resulted in longer and more explanatory comments that may assist code comprehension.
\end{answerbox}

%---------------------
% DISCUSSION
%---------------------
\section{Discussion}
\label{sec:disc}

We summarize the results of our experiment and discuss potential implications for research and practice.

\subsection{Developers are not Necessarily Prompting Experts}

Our study indicates that developers are generally neither skilled (RQ1) nor inclined (RQ2) to effectively prompt an \gls{llm} in a way that it generates concise, readable code documentation. 
This partly aligns with previous studies showing that optimized, predefined prompts outperform ad-hoc prompts \cite{ahmed-ase-2022}. 
But it also partly contradicts the underlying assumption that effective optimized prompts, such as few-shot prompts, are trivial and can intuitively be created by developers in their daily work. 
We observed in our experiment that participants interacted with the provided tools intuitively via natural language queries.
However, they were often disappointed by the results as the tools provided outputs that were aligned with their unspecific prompts.
In fact, we even observed that the generated documentation can partly be misleading during comprehending code. 

The lack of prompting skills among developers is unsurprising, as this is not a standard skill taught in universities or commonly practiced in software development.
Therefore, researchers should explore ways to learn effective prompting and to intuitively support conversations with \glspl{llm}, particularly in domains like software engineering where the \gls{llm} outcome has an impact on other users.
One approach is to compile a catalog of evaluated prompt templates for different well-defined tasks (e.g., documentation of Python functions).
Recently, Torricelli et al.~also found such templates beneficial for \gls{llm} users \cite{torricelli-websci-2024}.
Tool vendors could integrate the templates into their tools.
Each template could include instructions for the \gls{llm} to ask clarifying and contextual questions, e.g., related to the expected documentation style \cite{maalej-tse-2013}.
Additionally, educators could use these prompt templates, together with underlying ideas and hints, to teach prompting skills in schools and universities.
Such templates should only provide guidance and should not hinder human creativity in asking out-of-the-box and follow-up questions to an \gls{llm}-powered tool.

\subsection{Generating Documentation With LLMs is an Iterative Task}

Our results show that the participants rated the quality of documentation generated via the predefined prompt as high.
However, many were not completely satisfied with the documentation quality after this initial iteration.
The first iteration helped to comprehend the code, formulate the initial comment version, and gather ideas to alter and extend the documentation during further iterations.
This indicates that generating code documentation is rather an iterative task, not to be fully automated without the feedback of developers.

To better understand and assist this task, additional research should be conducted to identify and satisfy the preferences of code documentation providers and consumers \cite{maalej-tse-2013, maalej-pakm-2008, maalej-botse-2023}. 
Personal aspects include the depth of explanations, tone of the text, and lyrical style. 
Project aspects include documentation style used \cite{maalej-tse-2013} and knowledge needs of code users.
This research can help vendors of \gls{ai}-powered code documentation tools align the generated comments with the requirements of their users.
Thus, vendors can focus on applying and optimizing predefined prompts, which participants in our experiment preferred, while also offering additional iterations to customize the output based on preferences and needs.
Furthermore, generating multiple documentation versions at once might also stimulate developer creativity in exploring possibilities to document code \cite{liang-icse-2024}.

\subsection{Evaluating Code Documentation Quality Requires Human-Centered Quality Metrics}

As Hu et al.~\cite{hu-tosem-2022} pointed out, current metrics used to assess the quality of \gls{ai}-generated content do not align with human evaluation regarding code documentation generation.
For instance, the usefulness of generated content \cite{maalej-botse-2023} should be crucial in practice but is barely evaluated in common \gls{llm} benchmarks. 
Our results suggest that the assessment of code documentation quality depends on multiple factors \cite{tang-msr-2023}, such as the preferences of documentation consumers, the ratio between documentation conciseness and code complexity, and the availability of certain information and style that developers expect to be included in the documentation.
Current metrics for automatically assessing the quality of generated texts do not incorporate these aspects.
Hence, while automatically generating code documentation is fairly straightforward with \glspl{llm}, an expert assessment is still required to ensure good quality, e.g., during a code review.
It remains unclear how far this assessment can be fully automated, e.g., by other agents or other \glspl{llm}.
Future research should focus on creating and testing quality metrics for generated documentation through documentation analyses, expert surveys, and developer and benchmark studies.

\subsection{Better Code Comprehension Requires Support by Personalized LLM-Generated Explanations On The Fly}

The results of our code comprehension tasks show that \glspl{llm} can support developers in understanding code.
Especially, the longer explanations provided by the ad-hoc prompt tool in response to vague prompts helped participants understand the given code function.
However, such explanations can also mislead developers if the generated output does not consider the information needs \cite{maalej-tosem-2014} or provide the particular knowledge nuggets needed for the comprehension task \cite{fucci-fse-2019}.

Therefore, similar to the work of Nam et al.~\cite{nam-icse-2024}, we propose that vendors of \gls{ide}-integrated code comprehension tools should offer multiple code-related aspects that tool users can explore during the comprehension process.
Examples include the concepts mentioned in the code or the typical code usage \cite{nam-icse-2024}, but could also extend to explaining the code in a personalized way.
Researchers should survey developers to identify additional aspects and test their relevance in practice.

\subsection{Practitioners May Have Specific LLM Requirements}

Participants in our study expressed multiple requirements for an \gls{llm}-powered documentation generation tool.
For example, participants expressed concerns regarding security risks associated with using closed-source, externally hosted models and highlighted the extended time required for generating comments via the \gls{llm} model we used.

To address these issues, tool vendors should support a variety of \glspl{llm} within a single tool, including smaller, open-source models, to better meet the needs of organizations operating with sensitive data and under specific constraints.
One of the research challenges is that different models may respond variably to the same prompt, impacting the consistency and reliability of generated outputs.

%---------------------
% THREATS
%---------------------
\section{Threats to Validity}
\label{sec:threats}

\subsection{Construct Validity}
\label{sec:threats_construct}
% "Construct validity concerns generalizing the result of the experiment to the concept or theory behind the experiment." (Wohlin-Springer-2012)
% Design threats: Inadequate preoperational explanation of constructs, bias to due to mono-operation or mono-methdod that does not represent the full construct, confounding constructs, multiple construct levels
% Social threats: participants guess the hypothesis, participants are afraid to be evaluated (evaluation apprehension), explicit expectations from experimenter
(T01) Our study exclusively employed human evaluation techniques to assess code documentation quality, introducing a mono-method bias \cite{wohlin-springer-2012}.
Although automated metrics are used in other studies, we opted for human evaluation due to its limited correlation with automated metrics \cite{hu-tosem-2022}, and considered it beneficial for a more comprehensive understanding.

\subsection{Internal Validity}
\label{sec:threats_internal}
% "Threats to internal validity are influences that can affect the independent variable with respect to causality, without the researcher’s knowledge." (Wohlin-Springer-2012) 
(T02) We unintentionally excluded the \textit{Understandable} scale from our \gls{ueq}, which may compromise the \textit{Perspicuity} category's accuracy. 
Remedying this, we calculated Perspicuity scores with and without the scale's median value, finding only marginal differences. 
Despite this omission, participant feedback indicated an adequate understanding of the tool.
(T03) Customizability of the predefined prompt could have positively influenced participant satisfaction and tool assessment. 
However, allowing customization could have affected the comparability of assessments among the predefined prompt group. 
(T04) Lack of significant differences in professional ratings may be attributed to the small sample size of ten professionals per group. 
Nevertheless, descriptive statistics reveal fewer differences among the professionals' ratings compared to students, likely due to their more targeted use of keywords in ad-hoc prompts.
(T05) A potential language barrier existed given the non-native English-speaking background of most participants. 
However, this was deemed a minor threat as all were able to produce coherent prompts in English.

\subsection{External Validity}
\label{sec:threats_external}
% "Threats to external validity are conditions that limit our ability to generalize the results of our experiment to industrial practice. There are three types of interactions with the treatment: people, place and time." (Wohlin-Springer-2012)
% Not representative sample, unrealistic setting, toy problems, history (e.g., experiment on safety-critical systems a few days after a big software-related crash -> people tend to answer differently than a few days before, or some weeks or months later)
(T06) Our controlled lab setup, where developers documented unfamiliar code, may not accurately represent typical documentation practices. 
However, this setting was intended to eliminate variability and enable statistical analysis. 
(T07) The professionals in our study predominantly came from \gls{euxfel}, with specific expertise in Python, which might influence the results due to familiarity with the experimental code. 
This was seen as potentially beneficial, providing insight into how domain knowledge of one experiment group might improve their prompt creation.
(T08) The findings' generalizability is limited, yet they corroborate and broaden the insights from similar studies that involved different methodologies.

\subsection{Conclusion Validity}
\label{sec:threats_conclusion}
% "Threats to the conclusion validity are concerned with issues that affect the ability to draw the correct conclusion about relations between the treatment and the outcome of an experiment." (Wohlin-Springer-2012)
% Examples: Low statistical power, violated assumptions of statistical tests, fishing for results, reliability of measures (incl. replicability of results), reliability of the treatment across the participants, confounding factors during experiment, heterogeneous subjects (bad for conclusion val., good for external val.)
(T09) The selection of utility functions as code candidates might limit the applicability of our results to more complex coding tasks.
We chose these functions based on a pilot study confirming the difficulty of comprehending challenging code under time constraints.
(T10) Variations in \gls{llm} outputs due to inherent randomness \cite{tian-arxiv-2023} may have influenced assessments in the predefined prompt group, though differences in outputs were marginal.
(T11) Dividing the experimental and control groups into professionals (experts) and students (newcomers) reduced overall sample sizes, weakening statistical power but allowing for a more detailed analysis of differing perceptions between these groups.
The separation between experts and newcomers was confirmed by the participants' self-perceived Python experience.
Hence, we believe our findings can be generalized to a broader population.
Nevertheless, developers with different expertise in conversational agents might behave differently.

%---------------------
% RELATED WORK
%---------------------
\section{Related Work}
\label{sec:rw}

\subsection{Prompting Skills of Practitioners}
\label{sec:rw_ps}

Prompt engineering has been pivotal since the introduction of few-shot prompting \cite{brown-neurips-2020}. 
Ahmed et al. demonstrated its effectiveness in code summarization \cite{ahmed-ase-2022, ahmed-icse-2024}.
They highlighted that applying a few-shot prompt can outperform traditional models like \textit{CodeBERT} \cite{feng-emnlp-2020}, while zero- and one-shot prompts were less effective, aligning with findings of Geng et al. \cite{geng-icse-2024}. 
Meanwhile, subsequent techniques, like \textit{chain-of-thought prompting} \cite{wei-neurips-2022} and \textit{active prompts} \cite{diao-arxiv-2024}, have emerged.

Prompt engineering remains a skill requiring practice \cite{oppenlaender-arxiv-2023}.
\gls{llm} providers and researchers have published guidelines and prompt pattern catalogs for effective \gls{llm} interaction across various domains, including healthcare \cite{white-arxiv-2023, openai-online-2024, mesko-jmir-2023, heston-ime-2023}.
Our study is among the first to examine if and how well developers utilize prompt-related knowledge in software engineering tasks.

\subsection{LLM-Powered Developer Assistants}
\label{sec:rw_use}

Recent studies on \gls{ai} programming assistants show that tools like \textit{GitHub Copilot} \cite{github-online-2024} boost productivity, reduce keystrokes, and assist in syntax recall \cite{liang-icse-2024, zhang-seke-2023}.
However, users desire tool enhancements for direct feedback, personalized output, and improved code context understanding \cite{liang-icse-2024}.
Currently, they often avoid these tools due to unmet requirements and lack of control.
For example, Copilot often fails to grasp instructions for code adjustments unless precisely specified \cite{wermelinger-sigcse-2023}.
However, tool shortcuts to optimize an initial prompt can also nudge users to explore fewer prompts and lead to less diverse output \cite{torricelli-websci-2024}.
We can confirm these findings with our study.
While our participants overall preferred the interaction with and the output of the predefined prompt tool, the more exploratory ad-hoc prompts often led to more extensive output that supported code comprehension.

\subsection{Code Documentation Generation in the IDE}
\label{sec:rw_doc_gen}

% Models
In the last decade, researchers have developed various methods for generating code documentation \cite{rai-tist-2022}, including template-based \cite{sridhara-ase-2010, mcburney-icpc-2014}, information retrieval \cite{haiduc-icse-2010}, and, more recently, deep learning techniques \cite{hu-icpc-2018, alon-lr-2019, leclair-icpc-2020, gao-tosem-2023, wan-ase-2018, feng-emnlp-2020, wang-arxiv-2020}.
Especially deep learning approaches trained on large datasets have shown high performance in code documentation generation and enabled better scalability and more complete results than the other techniques \cite{rai-tist-2022}.
% Tools
We found many available \gls{ide} extensions that use these \gls{ai} models to facilitate developer workflows.
However, they were conceptualized as general-purpose tools and lacked code documentation features.
In the academic realm, the \gls{vscode} extension \textit{Divinator} was the most relevant tool to our project \cite{durelli-workshop-2022}. 
It provides short code summaries in multiple programming languages but has limited performance and interactivity.
With our study, we aimed to counter this lack of \gls{ide} features related to code documentation generation.

\subsection{Evaluating Documentation Quality}
\label{sec:rw_eval}

Empirical research on software documentation quality is an active field that focuses on various artifacts, like \gls{api} reference documentation \cite{maalej-tse-2013} or README files \cite{puhlfurss-icsme-2022}, and the perspectives of documentation writers \cite{aghajani-icse-2019}.
Studies on the evaluation of \gls{ai}-generated documentation usually focus on automated metrics like \textit{BLEU}, \textit{ROUGE}, and \textit{METEOR} \cite{aljumah-as-2022, gao-icpc-2022, gao-tosem-2023, sun-arxiv-2023, guo-tosem-2023}.
Hu et al.~compared such automated metrics with human evaluations on six documentation quality dimensions \cite{hu-tosem-2022} and found that automated metrics often misalign with human judgment \cite{hu-tosem-2022}. 
Consequently, our study benefited from their human-centered metric by capturing nuanced aspects in the generated code documentation that automated approaches may have missed.

%---------------------
% CONCLUSION
%---------------------
\section{Conclusion}
\label{sec:conc}

% Motivation
Recently, many \glspl{llm} have emerged, offering the potential to automate developer tasks, including creating code documentation.
However, it remained unclear how effectively developers could prompt \glspl{llm} to generate useful documentation.
% Methodology
We studied the interactions of developers with and their perception of \gls{llm}-powered \gls{ide} extensions during a controlled experiment with professionals and computer science students.
To generate documentation for two Python functions, the experimental group freely prompted an \gls{llm} and the control group applied a predefined few-shot prompt.
% Results
Our results revealed that students, who had relatively low Python experience, preferred the guidance of the few-shot prompt over ad-hoc prompting.
They rated the documentation generated by predefined prompts significantly higher in quality, particularly regarding readability, conciseness, usefulness, and helpfulness.
Professionals were more adept than students at including Python-specific keywords in their ad-hoc prompts, resulting in the generation of higher-quality documentation.
Consequently, they enjoyed the flexibility of ad-hoc prompting more than students, even if they did not apply prompt engineering techniques.
Overall, both types of \gls{llm} interactions improved the code comprehension of the study participants, but the participants often viewed the generated documentation as an intermediate result that needed iterative improvement.

We hope our findings encourage researchers to replicate this study \cite{kruse-zenodo-2024} in diverse settings, aiming to improve developers' prompting skills and \gls{ai}-powered tool support.

\section*{Acknowledgement}
We thank all participants of our experiment.
We acknowledge the support by DASHH (Data Science in Hamburg - Helmholtz Graduate School for the Structure of Matter) with the Grant-No. HIDSS-0002.

\bibliographystyle{IEEEtran}
\bibliography{bibliography}

\begin{thebibliography}{10}
\providecommand{\url}[1]{#1}
\csname url@samestyle\endcsname
\providecommand{\newblock}{\relax}
\providecommand{\bibinfo}[2]{#2}
\providecommand{\BIBentrySTDinterwordspacing}{\spaceskip=0pt\relax}
\providecommand{\BIBentryALTinterwordstretchfactor}{4}
\providecommand{\BIBentryALTinterwordspacing}{\spaceskip=\fontdimen2\font plus
\BIBentryALTinterwordstretchfactor\fontdimen3\font minus
  \fontdimen4\font\relax}
\providecommand{\BIBforeignlanguage}[2]{{%
\expandafter\ifx\csname l@#1\endcsname\relax
\typeout{** WARNING: IEEEtran.bst: No hyphenation pattern has been}%
\typeout{** loaded for the language `#1'. Using the pattern for}%
\typeout{** the default language instead.}%
\else
\language=\csname l@#1\endcsname
\fi
#2}}
\providecommand{\BIBdecl}{\relax}
\BIBdecl

\bibitem{aghajani-icse-2019}
E.~Aghajani, C.~Nagy, O.~L. Vega-Márquez, M.~Linares-Vásquez, L.~Moreno,
  G.~Bavota, and M.~Lanza, ``Software documentation issues unveiled,'' in
  \emph{2019 IEEE/ACM 41st International Conference on Software
  Engineering}.\hskip 1em plus 0.5em minus 0.4em\relax New York, NY, USA: IEEE,
  2019, pp. 1199--1210.

\bibitem{roehm-icse-2012}
T.~Roehm, R.~Tiarks, R.~Koschke, and W.~Maalej, ``How do professional
  developers comprehend software?'' in \emph{2012 34th International Conference
  on Software Engineering}.\hskip 1em plus 0.5em minus 0.4em\relax New York,
  NY, USA: IEEE, 2012, pp. 255--265.

\bibitem{steinmacher-software-2019}
I.~Steinmacher, C.~Treude, and M.~A. Gerosa, ``Let me in: Guidelines for the
  successful onboarding of newcomers to open source projects,'' \emph{IEEE
  Software}, vol.~36, no.~4, pp. 41--49, 2019.

\bibitem{stanik-icsme-2018}
C.~Stanik, L.~Montgomery, D.~Martens, D.~Fucci, and W.~Maalej, ``A simple
  {NLP}-based approach to support onboarding and retention in open source
  communities,'' in \emph{2018 IEEE International Conference on Software
  Maintenance and Evolution}, 2018, pp. 172--182.

\bibitem{zampetti-emse-2021}
F.~Zampetti, G.~Fucci, A.~Serebrenik, and M.~Di~Penta, ``Self-admitted
  technical debt practices: A comparison between industry and open-source,''
  \emph{Empirical Software Engineering}, vol.~26, no.~6, pp. 1--32, 2021.

\bibitem{aghajani-icse-2020}
E.~Aghajani, C.~Nagy, M.~Linares-V\'{a}squez, L.~Moreno, G.~Bavota, M.~Lanza,
  and D.~C. Shepherd, ``Software documentation: The practitioners'
  perspective,'' in \emph{Proceedings of the ACM/IEEE 42nd International
  Conference on Software Engineering}.\hskip 1em plus 0.5em minus 0.4em\relax
  New York, NY, USA: Association for Computing Machinery, 2020, p. 590–601.

\bibitem{rai-tist-2022}
S.~Rai, R.~C. Belwal, and A.~Gupta, ``A review on source code documentation,''
  \emph{ACM Transactions on Intelligent Systems and Technology}, vol.~13,
  no.~5, Jun. 2022.

\bibitem{bubeck-arxiv-2023}
S.~Bubeck, V.~Chandrasekaran, R.~Eldan, J.~Gehrke, E.~Horvitz, E.~Kamar,
  P.~Lee, Y.~T. Lee, Y.~Li, S.~Lundberg, H.~Nori, H.~Palangi, M.~T. Ribeiro,
  and Y.~Zhang, ``Sparks of artificial general intelligence: Early experiments
  with {GPT}-4,'' 2023.

\bibitem{ebert-software-2023}
C.~Ebert and P.~Louridas, ``Generative {AI} for software practitioners,''
  \emph{IEEE Software}, vol.~40, no.~4, pp. 30--38, 2023.

\bibitem{tian-arxiv-2023}
H.~Tian, W.~Lu, T.~O. Li, X.~Tang, S.-C. Cheung, J.~Klein, and T.~F.
  Bissyandé, ``Is {ChatGPT} the ultimate programming assistant -- how far is
  it?'' 2023.

\bibitem{wermelinger-sigcse-2023}
M.~Wermelinger, ``Using {GitHub Copilot} to solve simple programming
  problems,'' in \emph{Proceedings of the 54th ACM Technical Symposium on
  Computer Science Education V. 1}.\hskip 1em plus 0.5em minus 0.4em\relax New
  York, NY, USA: Association for Computing Machinery, 2023, p. 172–178.

\bibitem{liu-compsurv-2023}
P.~Liu, W.~Yuan, J.~Fu, Z.~Jiang, H.~Hayashi, and G.~Neubig, ``Pre-train,
  prompt, and predict: A systematic survey of prompting methods in natural
  language processing,'' \emph{ACM Computing Surveys}, vol.~55, no.~9, Jan.
  2023.

\bibitem{schulhoff-arxiv-2024}
S.~Schulhoff, M.~Ilie, N.~Balepur, K.~Kahadze, A.~Liu, C.~Si, Y.~Li, A.~Gupta,
  H.~Han, S.~Schulhoff, P.~S. Dulepet, S.~Vidyadhara, D.~Ki, S.~Agrawal,
  C.~Pham, G.~Kroiz, F.~Li, H.~Tao, A.~Srivastava, H.~D. Costa, S.~Gupta, M.~L.
  Rogers, I.~Goncearenco, G.~Sarli, I.~Galynker, D.~Peskoff, M.~Carpuat,
  J.~White, S.~Anadkat, A.~Hoyle, and P.~Resnik, ``The prompt report: A
  systematic survey of prompting techniques,'' 2024.

\bibitem{white-arxiv-2023}
J.~White, Q.~Fu, S.~Hays, M.~Sandborn, C.~Olea, H.~Gilbert, A.~Elnashar,
  J.~Spencer-Smith, and D.~C. Schmidt, ``A prompt pattern catalog to enhance
  prompt engineering with {ChatGPT},'' 2023.

\bibitem{openai-online-2024}
\BIBentryALTinterwordspacing
OpenAI. (2024) Prompt engineering. OpenAI, L.L.C. [Online]. Available:
  \url{https://platform.openai.com/docs/guides/prompt-engineering}
\BIBentrySTDinterwordspacing

\bibitem{brown-neurips-2020}
T.~Brown, B.~Mann, N.~Ryder, M.~Subbiah, J.~D. Kaplan, P.~Dhariwal,
  A.~Neelakantan, P.~Shyam, G.~Sastry, A.~Askell, S.~Agarwal, A.~Herbert-Voss,
  G.~Krueger, T.~Henighan, R.~Child, A.~Ramesh, D.~Ziegler, J.~Wu, C.~Winter,
  C.~Hesse, M.~Chen, E.~Sigler, M.~Litwin, S.~Gray, B.~Chess, J.~Clark,
  C.~Berner, S.~McCandlish, A.~Radford, I.~Sutskever, and D.~Amodei, ``Language
  models are few-shot learners,'' in \emph{Advances in Neural Information
  Processing Systems}, H.~Larochelle, M.~Ranzato, R.~Hadsell, M.~Balcan, and
  H.~Lin, Eds., vol.~33.\hskip 1em plus 0.5em minus 0.4em\relax Red Hook, NY,
  USA: Curran Associates, Inc., 2020, pp. 1877--1901.

\bibitem{logan-acl-2022}
R.~"Logan~IV, I.~Balazevic, E.~Wallace, F.~Petroni, S.~Singh, and S.~Riedel,
  ``"cutting down on prompts and parameters: Simple few-shot learning with
  language models",'' in \emph{"Findings of the Association for Computational
  Linguistics: ACL 2022"}, S.~"Muresan, P.~Nakov, and A.~Villavicencio,
  Eds.\hskip 1em plus 0.5em minus 0.4em\relax "Dublin, Ireland": "Association
  for Computational Linguistics", May "2022", pp. "2824--2835".

\bibitem{ahmed-ase-2022}
T.~Ahmed and P.~Devanbu, ``Few-shot training {LLM}s for project-specific
  code-summarization,'' in \emph{Proceedings of the 37th IEEE/ACM International
  Conference on Automated Software Engineering}.\hskip 1em plus 0.5em minus
  0.4em\relax New York, NY, USA: Association for Computing Machinery, 2023.

\bibitem{hu-tosem-2022}
X.~Hu, Q.~Chen, H.~Wang, X.~Xia, D.~Lo, and T.~Zimmermann, ``Correlating
  automated and human evaluation of code documentation generation quality,''
  \emph{ACM Trans. Softw. Eng. Methodol.}, vol.~31, no.~4, Jul. 2022.

\bibitem{kruse-zenodo-2024}
\BIBentryALTinterwordspacing
H.-A. Kruse, T.~Puhlf{\"u}r{\ss}, , and W.~Maalej, ``Can developers prompt? a
  controlled experiment for code documentation generation [replication
  package],'' 2024. [Online]. Available:
  \url{https://zenodo.org/doi/10.5281/zenodo.13127237}
\BIBentrySTDinterwordspacing

\bibitem{wohlin-springer-2012}
C.~Wohlin, P.~Runeson, M.~H{\"o}st, M.~C. Ohlsson, B.~Regnell, and
  A.~Wessl{\'e}n, \emph{Experimentation in software engineering}.\hskip 1em
  plus 0.5em minus 0.4em\relax Berlin, Heidelberg: Springer, 2012.

\bibitem{alhadreti-chi-2019}
O.~Alhadreti and P.~Mayhew, ``Rethinking thinking aloud: A comparison of three
  think-aloud protocols,'' in \emph{Proceedings of the 2018 CHI Conference on
  Human Factors in Computing Systems}.\hskip 1em plus 0.5em minus 0.4em\relax
  New York, NY, USA: Association for Computing Machinery, 2018, p. 1–12.

\bibitem{kapil-apress-2019}
S.~Kapil, \emph{Clean {Python}}.\hskip 1em plus 0.5em minus 0.4em\relax
  Berkeley, CA, USA: Apress, 2019.

\bibitem{laugwitz-hci-2008}
B.~Laugwitz, T.~Held, and M.~Schrepp, ``Construction and evaluation of a user
  experience questionnaire,'' in \emph{HCI and Usability for Education and
  Work}, A.~Holzinger, Ed.\hskip 1em plus 0.5em minus 0.4em\relax Berlin,
  Heidelberg: Springer Berlin Heidelberg, 2008, pp. 63--76.

\bibitem{macefield-jus-2007}
R.~Macefield, ``Usability studies and the {Hawthorne} effect,'' \emph{J.
  Usability Studies}, vol.~2, no.~3, p. 145–154, May 2007.

\bibitem{euxfel-online-2024}
\BIBentryALTinterwordspacing
EuropeanXFEL. (2024) Karabo. GitHub. [Online]. Available:
  \url{https://github.com/European-XFEL/Karabo/}
\BIBentrySTDinterwordspacing

\bibitem{rani-jss-2023}
P.~Rani, A.~Blasi, N.~Stulova, S.~Panichella, A.~Gorla, and O.~Nierstrasz, ``A
  decade of code comment quality assessment: A systematic literature review,''
  \emph{Journal of Systems and Software}, vol. 195, p. 111515, 2023.

\bibitem{hou-arxiv-2023}
X.~Hou, Y.~Zhao, Y.~Liu, Z.~Yang, K.~Wang, L.~Li, X.~Luo, D.~Lo, J.~Grundy, and
  H.~Wang, ``Large language models for software engineering: A systematic
  literature review,'' 2023.

\bibitem{sun-arxiv-2023}
W.~Sun, C.~Fang, Y.~You, Y.~Miao, Y.~Liu, Y.~Li, G.~Deng, S.~Huang, Y.~Chen,
  Q.~Zhang, H.~Qian, Y.~Liu, and Z.~Chen, ``Automatic code summarization via
  {ChatGPT}: How far are we?'' 2023.

\bibitem{renyang-online-2024}
\BIBentryALTinterwordspacing
Z.~Renyang. (2023) {ChatGPT}. Microsoft. [Online]. Available:
  \url{https://marketplace.visualstudio.com/items?itemName=zhang-renyang.chat-gpt}
\BIBentrySTDinterwordspacing

\bibitem{microsoft-online-2024}
\BIBentryALTinterwordspacing
Microsoft. (2024) Your first extension. Microsoft. [Online]. Available:
  \url{https://code.visualstudio.com/api/get-started/your-first-extension}
\BIBentrySTDinterwordspacing

\bibitem{redevtools-online-2024}
\BIBentryALTinterwordspacing
Re:DevTools. (2024) {Code Docs AI}. Microsoft. [Online]. Available:
  \url{https://marketplace.visualstudio.com/items?itemName=re-devtools.code-docs-ai}
\BIBentrySTDinterwordspacing

\bibitem{derrick-tqmp-2016}
B.~Derrick, D.~Toher, and P.~White, ``Why {Welch's} test is {Type I} error
  robust,'' \emph{The Quantitative Methods for Psychology}, vol.~12, no.~1, pp.
  30--38, 2016.

\bibitem{nachar-tqmp-2008}
N.~Nachar, ``The mann-whitney u: A test for assessing whether two independent
  samples come from the same distribution,'' \emph{Tutorials in Quantitative
  Methods for Psychology}, vol.~4, no.~1, pp. 13--20, 2008.

\bibitem{mohd-josmaa-2011}
N.~Mohd~Razali and B.~Yap, ``Power comparisons of {Shapiro-Wilk},
  {Kolmogorov-Smirnov}, {Lilliefors} and {Anderson-Darling} tests,''
  \emph{Journal of Statistical Modeling and Analytics}, vol.~2, no.~1, pp.
  21--33, 2011.

\bibitem{cohen-book-1988}
J.~Cohen, \emph{Statistical Power Analysis for the Behavioral Sciences}.\hskip
  1em plus 0.5em minus 0.4em\relax Routledge, 1988.

\bibitem{ueq-online-2024}
\BIBentryALTinterwordspacing
UEQ-team. (2024) User experience questionnaire. UEQ-team. [Online]. Available:
  \url{https://www.ueq-online.org}
\BIBentrySTDinterwordspacing

\bibitem{cascio-fm-2019}
M.~A. Cascio, E.~Lee, N.~Vaudrin, and D.~A. Freedman, ``A team-based approach
  to open coding: Considerations for creating intercoder consensus,''
  \emph{Field Methods}, vol.~31, no.~2, pp. 116--130, 2019.

\bibitem{torricelli-websci-2024}
M.~Torricelli, M.~Martino, A.~Baronchelli, and L.~M. Aiello, ``The role of
  interface design on prompt-mediated creativity in generative ai,'' in
  \emph{Proceedings of the 16th ACM Web Science Conference}.\hskip 1em plus
  0.5em minus 0.4em\relax New York, NY, USA: Association for Computing
  Machinery, 2024, p. 235–240.

\bibitem{maalej-tse-2013}
W.~Maalej and M.~P. Robillard, ``Patterns of knowledge in {API} reference
  documentation,'' \emph{IEEE Transactions on Software Engineering}, vol.~39,
  no.~9, pp. 1264--1282, 2013.

\bibitem{maalej-pakm-2008}
W.~"Maalej and H.-J. Happel, ``"a lightweight approach for knowledge sharing in
  distributed software teams",'' in \emph{"Practical Aspects of Knowledge
  Management"}, T.~"Yamaguchi, Ed.\hskip 1em plus 0.5em minus 0.4em\relax
  "Berlin, Heidelberg": "Springer Berlin Heidelberg", "2008", pp. "14--25".

\bibitem{maalej-botse-2023}
W.~Maalej, ``From {RSSE} to {BotSE}: Potentials and challenges revisited after
  15 years,'' in \emph{2023 IEEE/ACM 5th International Workshop on Bots in
  Software Engineering}, 2023, pp. 19--22.

\bibitem{liang-icse-2024}
J.~T. Liang, C.~Yang, and B.~A. Myers, ``A large-scale survey on the usability
  of {AI} programming assistants: Successes and challenges,'' in
  \emph{Proceedings of the 46th IEEE/ACM International Conference on Software
  Engineering}.\hskip 1em plus 0.5em minus 0.4em\relax New York, NY, USA:
  Association for Computing Machinery, 2024.

\bibitem{tang-msr-2023}
H.~Tang and S.~Nadi, ``Evaluating software documentation quality,'' in
  \emph{2023 IEEE/ACM 20th International Conference on Mining Software
  Repositories}, 2023, pp. 67--78.

\bibitem{maalej-tosem-2014}
W.~Maalej, R.~Tiarks, T.~Roehm, and R.~Koschke, ``On the comprehension of
  program comprehension,'' \emph{ACM Trans. Softw. Eng. Methodol.}, vol.~23,
  no.~4, sep 2014.

\bibitem{fucci-fse-2019}
D.~Fucci, A.~Mollaalizadehbahnemiri, and W.~Maalej, ``On using machine learning
  to identify knowledge in {API} reference documentation,'' in
  \emph{Proceedings of the 2019 27th ACM Joint Meeting on European Software
  Engineering Conference and Symposium on the Foundations of Software
  Engineering}.\hskip 1em plus 0.5em minus 0.4em\relax New York, NY, USA:
  Association for Computing Machinery, 2019, p. 109–119.

\bibitem{nam-icse-2024}
D.~Nam, A.~Macvean, V.~Hellendoorn, B.~Vasilescu, and B.~Myers, ``Using an
  {LLM} to help with code understanding,'' in \emph{Proceedings of the IEEE/ACM
  46th International Conference on Software Engineering}.\hskip 1em plus 0.5em
  minus 0.4em\relax New York, NY, USA: Association for Computing Machinery,
  2024.

\bibitem{ahmed-icse-2024}
T.~Ahmed, K.~S. Pai, P.~Devanbu, and E.~Barr, ``Automatic semantic augmentation
  of language model prompts (for code summarization),'' in \emph{Proceedings of
  the IEEE/ACM 46th International Conference on Software Engineering}.\hskip
  1em plus 0.5em minus 0.4em\relax New York, NY, USA: Association for Computing
  Machinery, 2024.

\bibitem{feng-emnlp-2020}
Z.~"Feng, D.~Guo, D.~Tang, N.~Duan, X.~Feng, M.~Gong, L.~Shou, B.~Qin, T.~Liu,
  D.~Jiang, and M.~Zhou, ``"{CodeBERT}: A pre-trained model for programming and
  natural languages",'' in \emph{"Findings of the Association for Computational
  Linguistics: EMNLP 2020"}, T.~"Cohn, Y.~He, and Y.~Liu, Eds.\hskip 1em plus
  0.5em minus 0.4em\relax "Online": "Association for Computational
  Linguistics", "Nov." "2020", pp. "1536--1547".

\bibitem{geng-icse-2024}
M.~Geng, S.~Wang, D.~Dong, H.~Wang, G.~Li, Z.~Jin, X.~Mao, and X.~Liao, ``Large
  language models are few-shot summarizers: Multi-intent comment generation via
  in-context learning,'' in \emph{Proceedings of the IEEE/ACM 46th
  International Conference on Software Engineering}.\hskip 1em plus 0.5em minus
  0.4em\relax New York, NY, USA: Association for Computing Machinery, 2024.

\bibitem{wei-neurips-2022}
J.~Wei, X.~Wang, D.~Schuurmans, M.~Bosma, b.~ichter, F.~Xia, E.~Chi, Q.~V. Le,
  and D.~Zhou, ``Chain-of-thought prompting elicits reasoning in large language
  models,'' in \emph{Advances in Neural Information Processing Systems},
  S.~Koyejo, S.~Mohamed, A.~Agarwal, D.~Belgrave, K.~Cho, and A.~Oh, Eds.,
  vol.~35.\hskip 1em plus 0.5em minus 0.4em\relax Curran Associates, Inc.,
  2022, pp. 24\,824--24\,837.

\bibitem{diao-arxiv-2024}
S.~Diao, P.~Wang, Y.~Lin, and T.~Zhang, ``Active prompting with
  chain-of-thought for large language models,'' 2024.

\bibitem{oppenlaender-arxiv-2023}
J.~Oppenlaender, R.~Linder, and J.~Silvennoinen, ``Prompting {AI} art: An
  investigation into the creative skill of prompt engineering,'' 2023.

\bibitem{mesko-jmir-2023}
B.~"Mesk{\'o}, ``"prompt engineering as an important emerging skill for medical
  professionals: Tutorial",'' \emph{"Journal of Medical Internet Research"},
  vol. "25", p. "e50638", "Oct." "2023".

\bibitem{heston-ime-2023}
T.~F. Heston and C.~Khun, ``Prompt engineering in medical education,''
  \emph{International Medical Education}, vol.~2, no.~3, pp. 198--205, 2023.

\bibitem{github-online-2024}
\BIBentryALTinterwordspacing
GitHub. (2024) {GitHub Copilot}. GitHub. [Online]. Available:
  \url{https://github.com/features/copilot}
\BIBentrySTDinterwordspacing

\bibitem{zhang-seke-2023}
B.~Zhang, P.~Liang, X.~Zhou, A.~Ahmad, and M.~Waseem, ``Demystifying practices,
  challenges and expected features of using {GitHub Copilot},''
  \emph{International Journal of Software Engineering and Knowledge
  Engineering}, vol.~33, no. 11n12, pp. 1653--1672, 2023.

\bibitem{sridhara-ase-2010}
G.~Sridhara, E.~Hill, D.~Muppaneni, L.~Pollock, and K.~Vijay-Shanker, ``Towards
  automatically generating summary comments for {Java} methods,'' in
  \emph{Proceedings of the 25th IEEE/ACM International Conference on Automated
  Software Engineering}.\hskip 1em plus 0.5em minus 0.4em\relax New York, NY,
  USA: Association for Computing Machinery, 2010, p. 43–52.

\bibitem{mcburney-icpc-2014}
P.~W. McBurney and C.~McMillan, ``Automatic documentation generation via source
  code summarization of method context,'' in \emph{Proceedings of the 22nd
  International Conference on Program Comprehension}.\hskip 1em plus 0.5em
  minus 0.4em\relax New York, NY, USA: Association for Computing Machinery,
  2014, p. 279–290.

\bibitem{haiduc-icse-2010}
S.~Haiduc, J.~Aponte, and A.~Marcus, ``Supporting program comprehension with
  source code summarization,'' in \emph{Proceedings of the 32nd ACM/IEEE
  International Conference on Software Engineering - Volume 2}.\hskip 1em plus
  0.5em minus 0.4em\relax New York, NY, USA: Association for Computing
  Machinery, 2010, p. 223–226.

\bibitem{hu-icpc-2018}
X.~Hu, G.~Li, X.~Xia, D.~Lo, and Z.~Jin, ``Deep code comment generation,'' in
  \emph{Proceedings of the 26th Conference on Program Comprehension}.\hskip 1em
  plus 0.5em minus 0.4em\relax New York, NY, USA: Association for Computing
  Machinery, 2018, p. 200–210.

\bibitem{alon-lr-2019}
U.~Alon, O.~Levy, and E.~Yahav, ``code2seq: Generating sequences from
  structured representations of code,'' in \emph{International Conference on
  Learning Representations}.\hskip 1em plus 0.5em minus 0.4em\relax Online:
  OpenReview, 2019.

\bibitem{leclair-icpc-2020}
A.~LeClair, S.~Haque, L.~Wu, and C.~McMillan, ``Improved code summarization via
  a graph neural network,'' in \emph{Proceedings of the 28th International
  Conference on Program Comprehension}.\hskip 1em plus 0.5em minus 0.4em\relax
  New York, NY, USA: Association for Computing Machinery, 2020, p. 184–195.

\bibitem{gao-tosem-2023}
S.~Gao, C.~Gao, Y.~He, J.~Zeng, L.~Nie, X.~Xia, and M.~Lyu, ``Code
  structure–guided transformer for source code summarization,'' \emph{ACM
  Trans. Softw. Eng. Methodol.}, vol.~32, no.~1, Feb. 2023.

\bibitem{wan-ase-2018}
Y.~Wan, Z.~Zhao, M.~Yang, G.~Xu, H.~Ying, J.~Wu, and P.~S. Yu, ``Improving
  automatic source code summarization via deep reinforcement learning,'' in
  \emph{Proceedings of the 33rd ACM/IEEE International Conference on Automated
  Software Engineering}.\hskip 1em plus 0.5em minus 0.4em\relax New York, NY,
  USA: Association for Computing Machinery, 2018, p. 397–407.

\bibitem{wang-arxiv-2020}
W.~Wang, Y.~Zhang, Z.~Zeng, and G.~Xu, ``{TranS}\^{}3: A transformer-based
  framework for unifying code summarization and code search,'' 2020.

\bibitem{durelli-workshop-2022}
R.~Durelli, V.~Durelli, R.~Bettio, D.~Dias, and A.~Goldman, ``Divinator: A
  visual studio code extension to source code summarization,'' in \emph{Anais
  do X Workshop de Visualização, Evolução e Manutenção de
  Software}.\hskip 1em plus 0.5em minus 0.4em\relax Porto Alegre, RS, Brasil:
  SBC, 2022, pp. 1--5.

\bibitem{puhlfurss-icsme-2022}
T.~Puhlfürß, L.~Montgomery, and W.~Maalej, ``An exploratory study of
  documentation strategies for product features in popular {GitHub} projects,''
  in \emph{2022 IEEE International Conference on Software Maintenance and
  Evolution}, 2022, pp. 379--383.

\bibitem{aljumah-as-2022}
S.~Aljumah and L.~Berriche, ``Bi-{LSTM}-based neural source code
  summarization,'' \emph{Applied Sciences}, vol.~12, no.~24, 2022.

\bibitem{gao-icpc-2022}
Y.~Gao and C.~Lyu, ``{M2TS}: Multi-scale multi-modal approach based on
  transformer for source code summarization,'' in \emph{Proceedings of the 30th
  IEEE/ACM International Conference on Program Comprehension}.\hskip 1em plus
  0.5em minus 0.4em\relax New York, NY, USA: Association for Computing
  Machinery, 2022, p. 24–35.

\bibitem{guo-tosem-2023}
H.~Guo, X.~Chen, Y.~Huang, Y.~Wang, X.~Ding, Z.~Zheng, X.~Zhou, and H.-N. Dai,
  ``Snippet comment generation based on code context expansion,'' \emph{ACM
  Trans. Softw. Eng. Methodol.}, vol.~33, no.~1, Nov. 2023.

\end{thebibliography}

\end{document}